\documentclass{article}
\usepackage{iclr2027_conference,times}

\usepackage{amsmath,amsfonts,bm}

\def\eqref#1{equation~\ref{#1}}

\def\1{\bm{1}}

\DeclareMathAlphabet{\mathsfit}{\encodingdefault}{\sfdefault}{m}{sl}
\SetMathAlphabet{\mathsfit}{bold}{\encodingdefault}{\sfdefault}{bx}{n}

\usepackage{amsmath,amssymb}
\usepackage{booktabs}
\usepackage{multirow}
\usepackage{colortbl}
\usepackage{graphicx}
\usepackage{placeins}
\usepackage{float}
\usepackage{wrapfig}

\usepackage{hyperref}
\usepackage{url}
\hypersetup{hidelinks}

\title{Learning to Explain While Planning: Rule-Aligned Diffusion Planning for Autonomous Driving}

\author{Jiaxi Ye \quad Chunji Lv \quad Guoren Wang \quad
Changsheng Li\thanks{Corresponding author.} \\
Beijing Institute of Technology}

\definecolor{bestblue}{RGB}{218,236,248}
\definecolor{refred}{RGB}{0,0,0}
\definecolor{citedarkyellow}{RGB}{0,0,0}
\newcommand{\best}[1]{\begingroup\setlength{\fboxsep}{1.2pt}\colorbox{bestblue}{\strut#1}\endgroup}
\newcommand{\radpfigref}[1]{Figure~\textcolor{refred}{\ref{#1}}}
\newcommand{\radptabref}[1]{Table~\textcolor{refred}{\ref{#1}}}
\newcommand{\radpappref}[1]{Appendix~\textcolor{refred}{\ref{#1}}}
\renewcommand{\citep}[1]{(\textcolor{citedarkyellow}{\citealp{#1}})}

\iclrfinalcopy
\begin{document}
\maketitle
\lhead{}
\renewcommand{\headrulewidth}{0pt}

\begin{abstract}
Diffusion planners exhibit strong capabilities in generating multimodal trajectories. However, existing methods primarily rely on expert demonstrations to fit trajectory distributions, learning statistical correlations among scenes, behaviors, and trajectories without explicitly modeling driving rules. In long-tail scenarios where expert data are scarce, the lack of behaviors to imitate may lead to trajectories that violate safety or compliance requirements. Moreover, their generation process lacks rule-level explanations, making it difficult to determine which rules drive trajectory adjustments, when they take effect, and how strongly they act, thereby limiting failure diagnosis, safety validation, and targeted improvement. To address these limitations, we propose the Rule-Aligned Diffusion Planner (RADP), which incorporates differentiable driving rules into the diffusion objective during training, turning rule knowledge into intrinsic behavioral principles beyond finite demonstrations. We further introduce Rule-Pressure Attribution (RPA), which constructs supervision signals from gradients of rule losses with respect to predicted trajectories and employs a lightweight attribution head to estimate the optimization pressure exerted by each rule online. To assess the closed-loop behavioral relevance of these attributions, we propose a temporal risk-alignment protocol that evaluates whether current rule pressures reflect corresponding risks during subsequent closed-loop execution. Experiments on nuPlan show that RADP improves closed-loop planning in challenging safety-critical scenarios, while RPA exhibits consistent temporal alignment with subsequent rule-specific risks, validating both intrinsic rule learning and rule-level interpretability.
\end{abstract}

\section{Introduction}

Diffusion models provide expressive representations of multimodal trajectory
distributions and have achieved strong closed-loop planning performance.
However, existing diffusion planners learn driving behavior mainly by imitating
expert trajectories. Requirements such as collision avoidance, lane compliance,
and speed-limit compliance are typically implicit in the data distribution
rather than explicit training objectives. When expert demonstrations do not
adequately cover complex interactions and long-tail states, a planner may learn
superficial behavior patterns without acquiring transferable rule constraints.
Although rewards or costs can guide diffusion sampling at inference time, such
rules remain external optimization signals and do not directly shape the
learned generative policy. Consequently, rule compliance remains decoupled from the 
model's  learned generative capability, potentially leading to inconsistent behavior 
under distribution shifts and safety-critical situations insufficiently represented in 
the training data.

Diffusion planners also lack faithful rule-level explanations of their planning decisions.
Existing interpretability techniques, including attention visualization and saliency mapping,
are predominantly post-hoc: they analyze a decision after it has been produced and identify
influential input features or internal activations. However, feature importance does not
directly reveal the semantic objectives that drive trajectory optimization, nor does it quantify
when and how strongly a particular driving rule acts on the generated trajectory. Such
post-hoc explanations are therefore weakly connected to the planner's actual optimization
mechanism and cannot naturally provide rule-level explanations together with trajectory
generation. In safety-critical autonomous driving, this limitation makes it difficult to
determine why a behavior was selected, localize the causes of planning failures, validate
safety properties, and perform targeted model improvement.

We therefore propose the Rule-Aligned Diffusion Planner (RADP), which jointly
optimizes the diffusion objective and differentiable driving rules during
training. The rules directly participate in learning the trajectory
distribution, without requiring additional rule guidance or trajectory
optimization at inference. Building on RADP, we introduce Rule-Pressure
Attribution (RPA): the gradient magnitude of each rule cost with respect to the
predicted trajectory represents its local optimization pressure and supervises
a lightweight attribution head. The planner can thus produce a trajectory and
its rule-level explanation within the same planning cycle.

To evaluate whether rule attributions reflect closed-loop driving conditions,
we introduce a temporal risk-alignment protocol that associates each rule’s
current attribution with its corresponding risk over a future rollout window
starting at the same planning frame. The protocol measures whether higher
rule-specific risks are accompanied by higher attribution pressures. Experiments on nuPlan show that RADP maintains broadly comparable overall
planning performance while improving collision and TTC metrics in challenging
safety-critical scenarios. The attribution head approximates the gradient
teacher, and its outputs exhibit stable temporal correspondence with subsequent
rule-specific risks.

Our contributions are summarized as follows:\par
\noindent\textbullet\enspace We propose a rule-aligned diffusion planning framework that integrates differentiable driving rules into training, enabling rule-consistent planning without inference-time rule-based refinement.\par

\noindent\textbullet\enspace We introduce online rule-pressure attribution, which distills trajectory-gradient-based rule pressures into a lightweight head, providing efficient rule-aware explanations alongside trajectory generation.\par

\noindent\textbullet\enspace We develop a temporal risk-alignment protocol that relates rule pressures to corresponding future risks in closed-loop rollouts, providing empirical validation of the effectiveness and behavioral relevance of our attributions.

\section{Related Work}

\paragraph{\textbf{Diffusion Planning.}}
Diffusion models have become a strong framework for multimodal trajectory
prediction and planning, extending denoising probabilistic modeling and
Transformer denoisers~\citep{ho2020ddpm,peebles2023dit} to sequential
decision-making and visuomotor policies~\citep{ajay2023decisiondiffuser,chi2023diffusionpolicy}.
MotionDiffuser supports controllable multi-agent
sampling through differentiable costs~\citep{jiang2023motiondiffuser}; Diffusion
Planner uses a Transformer-based diffusion architecture for joint prediction
and closed-loop planning~\citep{zheng2025diffusionplanner}; and DiffusionDrive
improves inference efficiency with multimodal anchors and a truncated diffusion
schedule~\citep{liao2025diffusiondrive}. Despite their planning performance,
these methods learn behavior mainly from demonstrations and do not explicitly
represent semantic rules for compliance reasoning or decision explanation.

\paragraph{\textbf{Rule and Constraint Integration in Learned Planning.}}

Beyond fitting expert trajectory distributions, prior work introduces
additional objectives into generative planning to improve safety and
controllability. Diffuser~\citep{janner2022planning} guides sampling toward
task-specific rewards; MotionDiffuser~\citep{jiang2023motiondiffuser} uses
differentiable costs that encode physical priors and user preferences;
Diffusion-ES~\citep{yang2024diffusiones} combines diffusion sampling with
evolutionary search to support black-box and non-differentiable objectives; and
guided conditional diffusion~\citep{zhong2023guided} applies semantic
constraints, including traffic rules, during denoising for controllable traffic
simulation. SafeDiffuser embeds control-barrier conditions into denoising to
enforce safety specifications~\citep{ames2017cbf,xiao2025safediffuser}, while diffusion
predictive control imposes explicit state and action constraints on sampled
controls~\citep{romer2025dpcc}; reward-based diffusion policy learning instead
connects policy scores to action-value gradients~\citep{psenka2024qscore}.
Despite different formulations, these objectives
primarily constrain or guide sampling after the diffusion model has been
learned, rather than making semantic driving rules part of policy training.

\paragraph{\textbf{Explainable Diffusion Planning.}}

Existing explainability methods use gradients, saliency, and attention
propagation to identify factors underlying neural-network
outputs~\citep{simonyan2013saliency,sundararajan2017integrated}
and related visual explanations~\citep{selvaraju2017gradcam,abnar2020quantifying,arrieta2020xai};
work on diffusion models further studies cross-attention, the denoising
process, and training-data contribution~\citep{tang2023daam,park2024explaining,dai2024training}.
Driving-specific approaches expose interpretable cost volumes and intermediate
predictions~\citep{zeng2019interpretable}, generate textual rationales grounded
by visual attention~\citep{kim2018textual}, or rank nearby agents using
interaction attention~\citep{hazard2022importance}. Interpretable trajectory
representations~\citep{ivanovic2021mats}, action-oriented
features~\citep{xiao2021action}, and language aligned with intermediate driving outputs
~\citep{ding2025hintad} provide complementary forms of transparency. Together with PlanT's
object-level attention~\citep{renz2023plant} and InterFuser's semantic
intermediate features~\citep{shao2023interfuser}, these methods explain input
objects, representations, or visual evidence. They do not directly quantify the
corrective pressure exerted by individual driving rules. We instead use
gradients of differentiable rules with respect to the trajectory as attribution
supervision for online rule-pressure estimation.

\section{Method}

\begin{figure*}[t]
    \centering
  \includegraphics[width=0.86\textwidth]{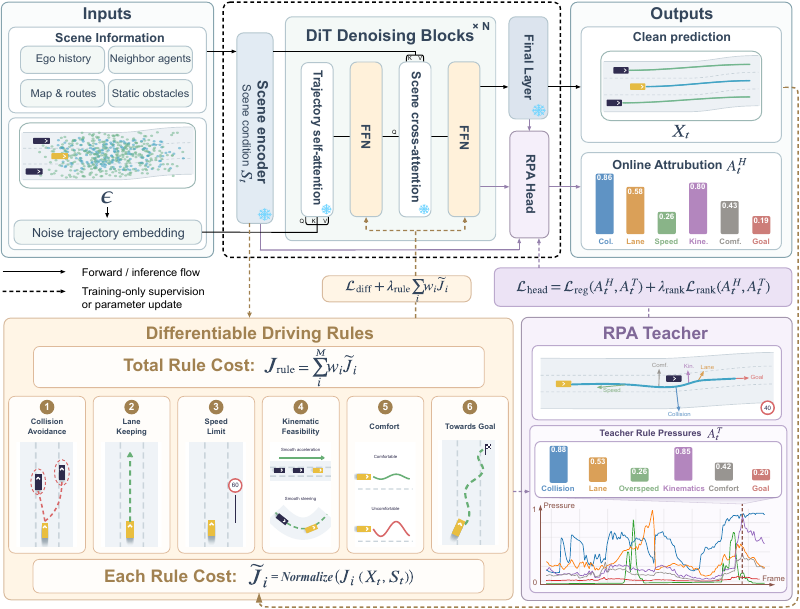}
    \caption{Overview of RADP.}
    \label{fig:radp_pipeline}
\end{figure*}

\subsection{Problem Setup and Method Overview}

Given the scene observation $S_t$ at planning time $t$, RADP produces a future
ego trajectory together with its online rule-level attribution within the same
planning cycle, as illustrated in \radpfigref{fig:radp_pipeline}. The observation
contains the histories of the ego vehicle and other traffic participants, map
elements, route information, and static obstacles; $\epsilon$ denotes the
diffusion noise. The figure further summarizes how differentiable rules first
adapt the planner and subsequently provide gradient-based supervision for the
attribution head, with these supervision paths used only during training. We
write the overall input--output relation as
\begin{equation}
    (X_t,A_t^H)=\mathcal{F}_{\theta,\phi}(S_t;\epsilon),
    \qquad
    X_t\in\mathbb{R}^{H\times d},\quad
    A_t^H\in\mathbb{R}_{\geq0}^{M},
    \label{eq:radp_io}
\end{equation}
where $X_t=P_\theta(S_t;\epsilon)$ is the predicted ego trajectory over a
horizon of $H$ steps, and $A_t^H$ is the non-negative rule-pressure vector
produced by the online attribution head $f_\phi$. We use $M=6$ channels:
physical collision avoidance, lane-boundary compliance, speed-limit
compliance, kinematic feasibility, comfort, and goal progress.

The method is trained in two stages.
Section~\ref{sec:differentiable-rules} defines the differentiable rule costs
\begin{equation}
    \mathcal{J}=\{J_i(X_t,S_t)\}_{i=1}^{M},
\end{equation}
and Section~\ref{sec:planner-adaptation} incorporates them into the diffusion
objective for lightweight planner adaptation. With the adapted planner frozen,
Section~\ref{sec:gradient-teacher} constructs gradient-based rule-pressure
targets, and Section~\ref{sec:online-head} distills them into an online
attribution head. At inference, only the planner and attribution head are
executed, without teacher gradients, rule guidance, or trajectory optimization.

\subsection{Differentiable Driving Rules}
\label{sec:differentiable-rules}

All rules operate on trajectories in physical coordinates at $0.1\,$s
resolution and are differentiable with respect to the predicted ego
trajectory. For rule $i$, we first define a trajectory-level cost
$J_i(X_t,S_t)$ that measures the overall degree to which trajectory $X_t$
violates that rule in scene $S_t$:
\begin{equation}
    J_i(X_t,S_t)
    =\operatorname{Agg}_{i,h,n}\!\left[
      m_{i,h,n}\,
      \varphi_{\sigma_i}\!\left(z_{i,h,n}(X_t,S_t)\right)
    \right].
    \label{eq:generic_rule_cost}
\end{equation}
Equation~\eqref{eq:generic_rule_cost} compactly maps local violations to a
trajectory-level cost. Reading it from the inside out,
$z_{i,h,n}(X_t,S_t)$ is the local physical violation of rule $i$ at future
step $h$ with respect to interaction entity $n$. The index $n$ is used only
for interactive rules such as collision avoidance and is omitted for rules
that depend solely on the ego state. We use the convention
$z_{i,h,n}>0$ for a violation and $z_{i,h,n}\leq0$ for rule satisfaction or a
remaining safety margin. For example, collision uses the required safety
distance minus the actual box separation, whereas speed-limit compliance uses
the predicted speed minus the permitted speed.

To convert this signed, physically dimensioned quantity into a non-negative
and differentiable local penalty, we use
\begin{equation}
    \varphi_{\sigma_i}(z)
    =\left[\frac{1}{\beta}
      \operatorname{softplus}\!\left(\frac{\beta z}{\sigma_i}\right)\right]^2,
    \qquad \beta=10.
    \label{eq:smooth_violation}
\end{equation}
Here $\sigma_i$ is the physical scale of rule $i$. The function approximates
$[\max(0,z/\sigma_i)]^2$: its value is close to zero when the rule is
satisfied and increases smoothly with violation severity. In
Eq.~\eqref{eq:generic_rule_cost}, $m_{i,h,n}$ is an applicability or risk gate
that masks invalid or irrelevant local constraints, while
$\operatorname{Agg}_i$ aggregates the valid time steps and interaction
entities into the scalar $J_i$. We instantiate six semantic channels covering
collision, lane, speed, kinematics, comfort, and goal progress, as summarized
in \radptabref{tab:driving_rules}. Their complete definitions are provided in
\radpappref{app:rules}.

Because the resulting rule costs have different typical numerical ranges, we
normalize them before combining them in the planner objective. For sample $b$,
\begin{equation}
    \widetilde{J}_i^{(b)}
    =\frac{J_i^{(b)}}{s_i^{(r)}},
    \label{eq:normalized_rule_cost}
\end{equation}
Here $s_i^{(r)}$ is a stop-gradient EMA scale maintained separately for rule
$i$ during training (EMA decay $0.99$, lower bounded by $10^{-3}$). This
normalization balances complete rule costs in joint training without changing
their physical definitions. The attribution teacher uses the gradient of the
unnormalized $J_i$.

\begin{table}[t]
\centering
\scriptsize
\setlength{\tabcolsep}{3.5pt}
\renewcommand{\arraystretch}{1.08}
\caption{Six rule channels for planning and attribution.}
\label{tab:driving_rules}
\begin{tabular}{p{0.15\linewidth}p{0.45\linewidth}p{0.31\linewidth}}
\toprule
\textbf{Rule channel} & \textbf{Violation signal} & \textbf{Purpose} \\
\midrule
Collision & Insufficient ego--agent clearance & Avoid imminent collisions \\
Lane & Route-corridor boundary violation & Maintain lane and route compliance \\
Speed & Speed above the legal limit & Enforce speed-limit compliance \\
Kinematics & Excess acceleration, braking, or steering & Ensure dynamic feasibility \\
Comfort & Excess jerk or control variation & Encourage smooth motion \\
Goal & Insufficient route progress & Preserve planning efficiency \\
\bottomrule
\end{tabular}
\renewcommand{\arraystretch}{1}
\end{table}

As an example, collision avoidance uses oriented ego boxes constructed from
the predicted rear-axle states and the nuPlan Pacifica geometry. Let
$d_{h,n}^{\mathrm{box}}$ be the signed box separation computed by the
separating-axis theorem. The collision violation is
\begin{equation}
    z_{h,n}^{\mathrm{col}}
    = d_{h,n}^{\mathrm{safe}}-d_{h,n}^{\mathrm{box}},
    \qquad
    d_{h,n}^{\mathrm{safe}}
    = d_0+t_{\mathrm{head}}c_{h,n}
      +\frac{c_{h,n}^{2}}{2a_{\mathrm{brake}}},
    \label{eq:collision_violation}
\end{equation}
where $c_{h,n}\geq0$ is the closing speed and $d_0=0.5\,\mathrm{m}$.
Thus, $z_{h,n}^{\mathrm{col}}>0$ indicates insufficient clearance. We use
$\sigma_{\mathrm{col}}=0.5\,\mathrm{m}$ and a detached gate based on distance,
closing speed, TTC, and violation severity to suppress irrelevant
interactions. The resulting penalties are aggregated into $J_{\mathrm{col}}$,
with other-agent trajectories treated as fixed context.

\subsection{Rule-Aligned Planner Training}
\label{sec:planner-adaptation}

We build on Diffusion Planner~\citep{zheng2025diffusionplanner}, which models
the joint future trajectories of the ego vehicle and surrounding agents with an
$x_0$-prediction objective. We retain its masked ego--neighbor reconstruction
loss $\mathcal{L}_{\mathrm{diff}}$ and augment it with differentiable rule
costs (full baseline formulation in \radpappref{app:diffusion-objective}):
\begin{equation}
\mathcal{L}_{\mathrm{RADP}}
=
\mathcal{L}_{\mathrm{diff}}
+
\lambda_{\mathrm{rule}}
\sum_{i=1}^{M} w_i \widetilde{J}_i,
\label{eq:planner_training}
\end{equation}
where $M=6$, $\widetilde{J}_i$ is the normalized cost of rule $i$, and $w_i$
controls its contribution to planner training. We set
$\lambda_{\mathrm{rule}}=0.004$ and use
$(w_{\mathrm{col}},w_{\mathrm{lane}},w_{\mathrm{speed}},w_{\mathrm{kin}},
w_{\mathrm{comfort}},w_{\mathrm{goal}})=(2.5,0.8,0.8,0.5,0.8,1.5)$.
Starting from the baseline checkpoint, we fine-tune lightweight LoRA adapters
in the decoder~\citep{hu2022lora}.

\subsection{Gradient-Pressure Teacher}
\label{sec:gradient-teacher}

\paragraph{\textbf{Rule-gradient teacher.}}
For a generated trajectory $X_t$, the gradient of a differentiable rule cost
indicates the first-order trajectory change required to reduce that cost. We
therefore define the teacher pressure for rule $i$ as the root-mean-square
magnitude of its trajectory gradient:
\begin{equation}
    g_i^T(t)
    =\left(
      \frac{1}{D}
      \left\|\nabla_{X_t}J_i(X_t,S_t)\right\|_2^2
    \right)^{1/2},
    \qquad D=Hd.
    \label{eq:rms_gradient}
\end{equation}
The resulting $g_i^T(t)$ is an independent, non-negative measure of the
trajectory's sensitivity to rule $i$: a larger value means that this rule
exerts stronger local pressure on the current plan.

\paragraph{\textbf{Scale-calibrated attribution.}}
Gradient magnitudes can differ systematically across rules. We estimate a
positive calibration scale $\kappa_i$ for each rule from the training split
and define the teacher attribution as
\begin{equation}
    A_i^T(t)
    =\log\left(1+\frac{g_i^T(t)}{\kappa_i+\varepsilon}\right).
    \label{eq:teacher}
\end{equation}
The logarithm attenuates rare extreme gradients while preserving their order.
The resulting $A_i^T(t)$ is a calibrated, rule-specific measure of local
trajectory sensitivity, which we use as supervision for the online attribution
head.

\subsection{Online Attribution Head}
\label{sec:online-head}

\begin{figure*}[t]
    \centering
  \includegraphics[width=\textwidth]{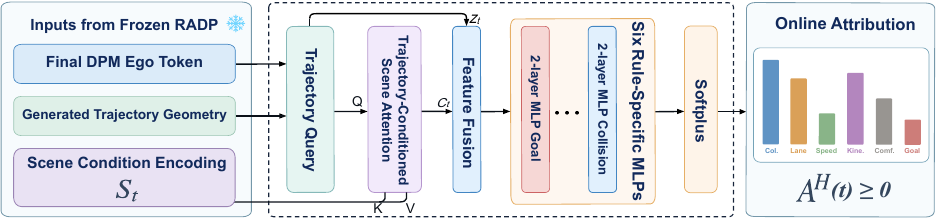}
    \caption{Online rule-pressure attribution head.}
    \label{fig:online-attribution-head}
\end{figure*}

Computing Eq.~\eqref{eq:teacher} for all rules requires repeated backward
passes. We amortize this computation with an attribution head $f_\phi$. Its base input
combines the ego token from the final DPM denoising call with geometric features
of the generated trajectory:
\begin{equation}
    z_t=[h_t^{\mathrm{DPM}};q(X_t)].
    \label{eq:head}
\end{equation}
The head also receives the scene condition encoding $S_t$, which jointly
contains surrounding-agent, map, route, and lane information. Using $z_t$ as
a query, it attends to the scene encoding and aggregates the context relevant
to the current generated trajectory:
\begin{equation}
    c_t=\operatorname{MHA}_{\mathrm{scene}}(z_t,S_t,S_t), \qquad
    A_i^H(t)=\operatorname{Softplus}\!\left(f_{\phi,i}([z_t;c_t])\right).
    \label{eq:token_head}
\end{equation}
Invalid tokens in the scene condition encoding are masked in attention. We
instantiate one independent two-layer MLP for each rule. All six MLPs receive
the shared representation $[z_t;c_t]$, but their output parameters are not
shared, producing independent non-negative pressures. The scene encoding lets
the head combine the current trajectory with agent interactions and road and
route constraints, rather than relying only on a compressed ego feature.
This computation is summarized in \radpfigref{fig:online-attribution-head}.

The planner parameters $\theta$ are frozen, and only $\phi$ is trained on
cached teacher targets. Our training objective combines
weighted pressure regression with a ranking term:
\begin{equation}
    \mathcal{L}_{\mathrm{head}}
    =\mathcal{L}_{\mathrm{reg}}(A^H,A^T)
    +\lambda_{\mathrm{rank}}\mathcal{L}_{\mathrm{rank}}.
    \label{eq:head_loss}
\end{equation}
We use a weighted Smooth-$L_1$ regression loss and a within-mini-batch
pairwise ranking loss, with additional weight on safety-critical rules and
high-pressure Teacher targets. The head is trained solely on cached gradient-teacher 
attributions. At inference, it reuses planner features in the same forward pass and does not alter the
generated trajectory; closed-loop rollout risks are reserved exclusively for
evaluating attribution quality.
\section{Experiments}

\subsection{Experimental Setup}

We evaluate RADP from three complementary perspectives. First, we assess
closed-loop planning performance on the nuPlan val14, test14-random, and
test14-hard reactive splits~\citep{caesar2021nuplan,karnchanachari2024nuplan}.
We compare RADP with Diffusion Planner across the overall score and all reported
metric components, and further compare the overall score with representative
learning-based planners. Non-reactive results are retained in
\radptabref{tab:planner-nr} in the appendix. Second, on synchronized closed-loop
planning frames, we evaluate how faithfully the online attribution head
approximates the gradient-based Teacher. Third, we examine whether each
rule-specific attribution is aligned with the corresponding driving condition
observed during closed-loop execution. We evaluate six attribution channels
against seven future-risk endpoints: each channel has one matched endpoint,
while the collision channel is additionally evaluated against TTC risk as a
complementary measure of imminent safety. Thus, TTC is an auxiliary evaluation
endpoint rather than a seventh Head output. Detailed risk endpoints and
statistical procedures are provided in Appendix~\textcolor{refred}{B}.

\subsection{Planner Performance}

\FloatBarrier
\begin{table}[H]
\centering
\caption{Closed-loop planning performance with reactive agents.}
\label{tab:planner}
\scriptsize
\setlength{\tabcolsep}{3.8pt}
\renewcommand{\arraystretch}{1.04}
\begin{tabular}{llccccccc}
\toprule
\textbf{Split} & \textbf{Planner} & \textbf{Score} & \textbf{Collision} & \textbf{TTC} & \textbf{Drivable} & \textbf{Progress} & \textbf{Speed} & \textbf{Comfort} \\
\midrule
\multirow{2}{*}{val14}
& Diffusion Planner & \best{82.70} & 92.98 & 87.84 & 97.85 & \best{96.60} & 98.29 & \best{89.45} \\
& RADP(ours) & 81.37 & \best{94.01} & \best{89.53} & \best{98.03} & 94.90 & \best{98.87} & 89.27 \\
\addlinespace[2pt]
\multirow{2}{*}{test14-random}
& Diffusion Planner & 82.82 & 94.06 & 89.66 & 98.08 & \best{97.70} & 97.07 & 85.44 \\
& RADP(ours) & \best{84.48} & \best{96.36} & \best{91.95} & 98.08 & 96.93 & \best{97.98} & \best{87.36} \\
\addlinespace[2pt]
\multirow{2}{*}{test14-hard}
& Diffusion Planner & 68.94 & 86.95 & 79.78 & 94.85 & \best{92.65} & 97.39 & 85.29 \\
& RADP(ours) & \best{70.62} & \best{92.10} & \best{83.46} & \best{95.59} & 90.07 & \best{98.29} & 85.29 \\
\bottomrule
\end{tabular}
\renewcommand{\arraystretch}{1}
\end{table}

\begin{table}[H]
\begin{minipage}[t]{0.38\textwidth}
\vspace{0pt}
As reported in \radptabref{tab:planner}, RADP improves the reactive score and
safety metrics on both test splits. On val14-r, safety also improves, but lower
progress reduces the aggregate score, revealing a safety--efficiency trade-off.
The broader comparison in \radptabref{tab:reported-planners} further shows that
RADP achieves the highest reported scores on test14-random and test14-hard while
remaining competitive on val14.
\end{minipage}\hfill
\begin{minipage}[t]{0.59\textwidth}
\vspace{0pt}
\centering
\caption{Literature-reported reactive nuPlan scores.}
\label{tab:reported-planners}
\scriptsize
\setlength{\tabcolsep}{2.8pt}
\renewcommand{\arraystretch}{1.04}
\begin{tabular}{@{}llccc@{}}
\toprule
\textbf{Type} & \textbf{Planner} & \textbf{Val14-R} & \textbf{Hard-R} & \textbf{Random-R} \\
\midrule
Expert & Log-replay & 80.32 & 68.80 & 75.86 \\
\midrule
\multirow{7}{*}{Learned}
& PDM-Open & 54.24 & 35.83 & 57.23 \\
& UrbanDriver & 64.11 & 49.95 & 67.15 \\
& GameFormer & 8.69 & 6.69 & 9.31 \\
& PlanTF & 76.95 & 61.61 & 79.58 \\
& PLUTO & 78.11 & 59.74 & 78.62 \\
& Diffusion Planner & \best{82.70} & 68.94 & 82.82 \\
& \textbf{RADP (ours)} & 81.37 & \best{70.62} & \best{84.48} \\
\bottomrule
\end{tabular}
\renewcommand{\arraystretch}{1}
\end{minipage}
\end{table}

\subsection{Attribution Evaluation}

We evaluate two complementary properties. On the same closed-loop planning
frame, scene state, and generated trajectory, macro Spearman, MAE, and Top-1
agreement respectively measure whether the Head preserves the Teacher's
pressure ordering, numerical values, and dominant-rule decision. Complete
definitions are provided in \radpappref{app:fidelity-metrics}.

Attribution $A_i(t_k)$ describes the pressure exerted by rule $i$ on the
trajectory generated at planning time $t_k$. To determine whether this pressure
corresponds to actual driving conditions, we construct future-risk targets from
the \emph{same closed-loop rollout}. Let $v_i^{\mathrm{rollout}}(\tau)$ denote
the instantaneous severity of rule-$i$ risk observed at execution time $\tau$.
The future risk associated with the current planning frame is
\begin{equation}
    R_i(t_k)
    =\operatorname{Agg}_{\tau\in[t_k,t_k+H_i]}
      v_i^{\mathrm{rollout}}(\tau),
    \label{eq:future_risk}
\end{equation}
where $H_i$ is a rule-specific horizon and $\operatorname{Agg}$ reduces
frame-wise severity within the window to a continuous risk value. Collision and
TTC use a two-second window because they describe imminent hazards. Lane
boundary, speed limit, kinematics, comfort, and goal progress use an eight-second
window. Rule-specific severity functions and aggregation operators are fixed,
together with all horizons and thresholds, before inspecting attribution
results and are shared by Teacher and Head. Hence, $R_i(t_k)$ is an independent
closed-loop evaluation target, not a training label.

We then test behavioral validity against $R_i(t_k)$. Scenario-wise temporal
Spearman measures whether attribution and subsequent rule-matched risk rise and
fall consistently; AUPRC evaluates retrieval of future high-risk frames across
thresholds; and Lift@10\% measures risk-event enrichment among the highest-
attribution 10\% of frames. Within-scenario temporal shuffling serves as a
control. Complete thresholds, eligibility criteria, formulas, and bootstrap
procedures are provided in \radpappref{app:temporal-metrics}.

\subsubsection{Closed-Loop Head--Teacher Fidelity}

On synchronized reactive frames, Table 4 shows that the Head consistently tracks the Teacher in pressure ranking, absolute magnitude, and dominant-rule identification, while both remain positively aligned with subsequent rule-matched risks.

\begin{table}[H]
\centering
\caption{Head--Teacher fidelity and future-risk alignment in reactive
closed-loop simulation.}
\label{tab:fidelity}
\footnotesize
\renewcommand{\arraystretch}{1.06}
\setlength{\tabcolsep}{3.8pt}
\resizebox{\textwidth}{!}{%
\begin{tabular}{lccccc}
\toprule
\multirow{2}{*}{\textbf{Reactive split}} & \multicolumn{3}{c}{\textbf{Head--Teacher fidelity}} & \multicolumn{2}{c}{\textbf{Future-risk alignment}} \\
\cmidrule(lr){2-4}\cmidrule(lr){5-6}
& \textbf{Macro Spearman $\uparrow$} & \textbf{MAE $\downarrow$} & \textbf{Top-1 $\uparrow$} & \textbf{Teacher $\rho_{\mathrm{macro}}$ $\uparrow$} & \textbf{Head $\rho_{\mathrm{macro}}$ $\uparrow$} \\
\midrule
val14 & 0.536 & 0.216 & 70.6\% & 0.235 & 0.207 \\
test14-random & 0.570 & 0.200 & 71.7\% & 0.199 & 0.177 \\
test14-hard & 0.554 & 0.238 & 71.9\% & 0.221 & 0.192 \\
\bottomrule
\end{tabular}
}
\renewcommand{\arraystretch}{1}
\end{table}

\subsubsection{Temporal Risk Alignment}

\radptabref{tab:temporal-val} details the alignment of Teacher and Head pressures with matched driving conditions on val14-r, where collision pressure is evaluated against both physical contact and auxiliary TTC risk. Complete confidence intervals are provided in the appendix.

\begin{table}[H]
\centering
\caption{Frame-level future-risk alignment on val14-r.}
\label{tab:temporal-val}
\scriptsize
\renewcommand{\arraystretch}{1.06}
\setlength{\tabcolsep}{2.8pt}
\resizebox{\textwidth}{!}{%
\begin{tabular}{lllcccccccc}
\toprule
\multirow{2}{*}{\textbf{Channel}} & \multirow{2}{*}{\textbf{Risk endpoint}} & \multirow{2}{*}{\textbf{Pos. rate}} & \multicolumn{2}{c}{\textbf{AUPRC $\uparrow$}} & \multicolumn{2}{c}{\textbf{Lift@10\% $\uparrow$}} & \multicolumn{2}{c}{\textbf{Temporal $\rho$ $\uparrow$}} & \multicolumn{2}{c}{\textbf{Shuffled $\rho$}} \\
\cmidrule(lr){4-5}\cmidrule(lr){6-7}\cmidrule(lr){8-9}\cmidrule(lr){10-11}
& & & \textbf{Teacher} & \textbf{Head} & \textbf{Teacher} & \textbf{Head} & \textbf{Teacher} & \textbf{Head} & \textbf{Teacher} & \textbf{Head} \\
\midrule
\multirow{2}{*}{Collision} & Physical collision & 2.37\% & 0.540 & 0.514 & 1.823 & 1.598 & 0.234 & 0.160 & 0.001 & $-0.002$ \\
& TTC risk (aux.) & 4.11\% & 0.555 & 0.489 & \best{1.926} & \best{1.610} & 0.237 & 0.158 & $-0.000$ & $-0.000$ \\
\addlinespace[1pt]
Lane & Lane-boundary risk & 49.31\% & \best{0.911} & \best{0.840} & 1.518 & 1.340 & 0.314 & 0.234 & 0.001 & 0.000 \\
Speed & Overspeed risk & 6.87\% & 0.779 & 0.761 & 1.890 & 1.895 & 0.303 & 0.297 & $-0.001$ & $-0.001$ \\
Kinematics & Kinematic risk & 0.013\% & 0.167 & 0.250 & 0.000 & 0.000 & 0.112 & 0.159 & $-0.000$ & 0.000 \\
Comfort & Comfort risk & 0.102\% & 0.376 & 0.371 & 1.750 & 1.750 & 0.102 & 0.122 & 0.000 & $-0.000$ \\
Goal progress & Route-progress deficit & 47.06\% & 0.850 & 0.798 & 1.804 & 1.626 & \best{0.343} & \best{0.316} & 0.005 & 0.027 \\
\bottomrule
\end{tabular}
}
\renewcommand{\arraystretch}{1}
\end{table}

For the five endpoints with sufficient events, Head AUPRC clearly exceeds the positive-rate baseline and Lift@10\% indicates risk enrichment among high-pressure frames. Specifically, AUPRC should be compared with the positive event rate, Lift@10\% with 1, and temporal correlation with 0. Across rule categories, the observed positive temporal correlations, together with near-zero correlations after temporal shuffling, further support a meaningful temporal association between predicted rule pressures and subsequent rule-specific risks.

\subsection{Efficiency}

\begin{table}[H]
\begin{minipage}[t]{0.39\textwidth}
\vspace{0pt}
As reported in \radptabref{tab:latency}, adding the attribution Head leaves
end-to-end planning latency essentially unchanged: planner-only and
planner-plus-Head inference require 251.05 and 245.50 ms per step respectively. Thus, online attribution introduces no
measurable end-to-end overhead in this evaluation.
\end{minipage}\hfill
\begin{minipage}[t]{0.58\textwidth}
\vspace{0pt}
\centering
\caption{Mean inference latency on an RTX~4080.}
\label{tab:latency}
\scriptsize
\setlength{\tabcolsep}{2.8pt}
\begin{tabular}{@{}lrr@{}}
\toprule
\textbf{Method} & \textbf{Latency (ms)} & \textbf{Overhead} \\
\midrule
Planner only & 251.05 & --- \\
Planner + Head & 245.50 & $-1.64\%$ \\
Planner + exact Teacher & 465.29 & $+89.18\%$ \\
Planner + gradient guidance & 571.10 & $+130.76\%$ \\
\midrule
Exact Teacher on cached states & 62.51 & $1.00\times$ \\
Head on cached states & \best{8.18} & \best{$7.65\times$ faster} \\
\bottomrule
\end{tabular}
\end{minipage}
\end{table}

In contrast, computing the exact Teacher online increases latency to 465.29
ms, and inference-time collision-gradient guidance requires 571.10 ms. The
cached-state comparison excludes trajectory generation and measures only the
additional attribution computation on the same stored planner states. Under
this setting, the Head requires 8.18 ms compared with 62.51 ms for the exact
Teacher, yielding a \(7.65\times\) attribution speedup.

\subsection{Qualitative Analysis}

\begin{figure}[H]
\centering
\begin{minipage}[c]{0.62\textwidth}
    \vspace{0pt}
    \centering
    \includegraphics[width=\linewidth]{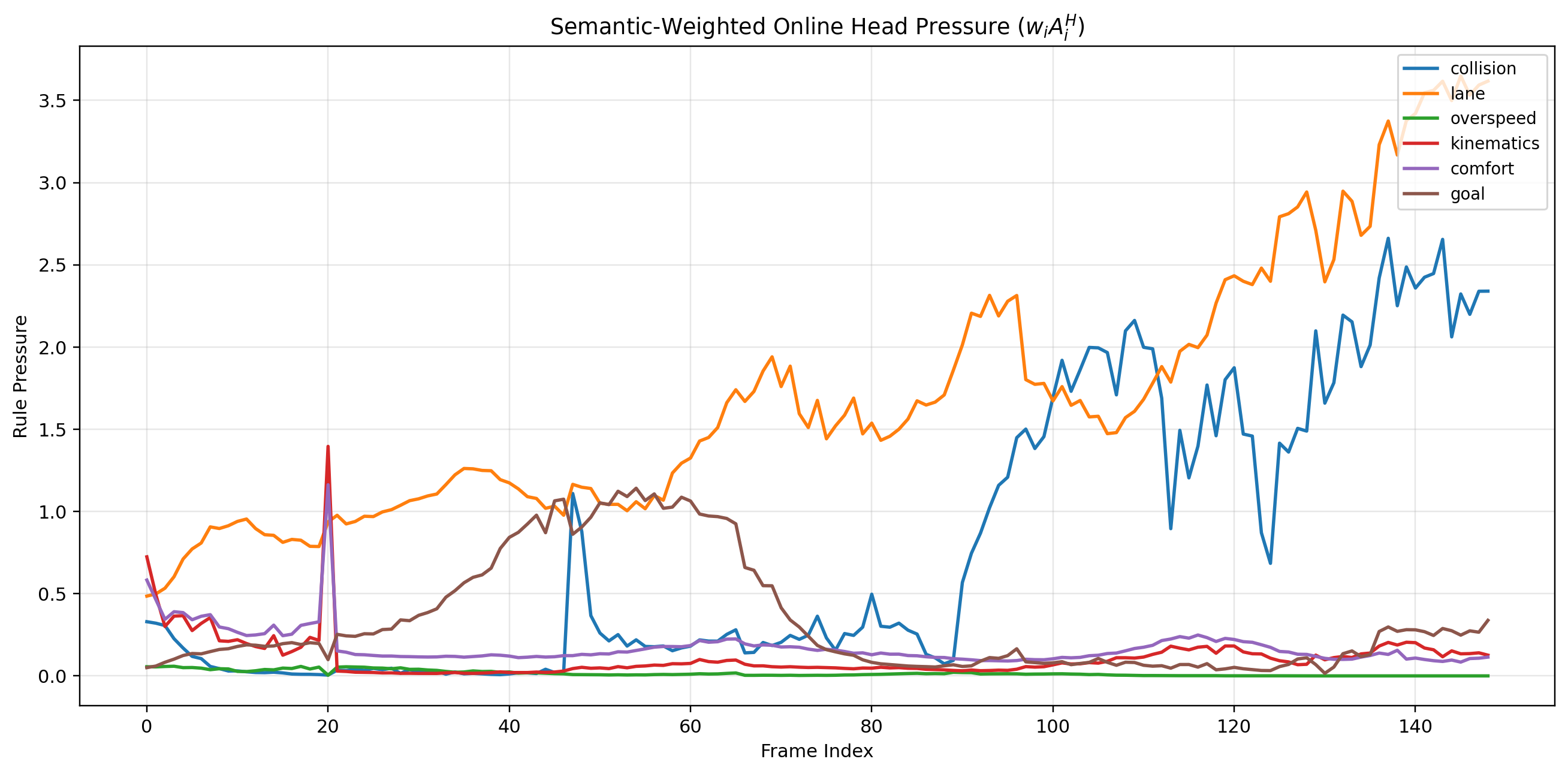}
    \par\vspace{0.5mm}
    \small (a) online rule pressure $A_i^H$
\end{minipage}\hfill
\begin{minipage}[c]{0.34\textwidth}
    \centering
    \includegraphics[width=\linewidth]{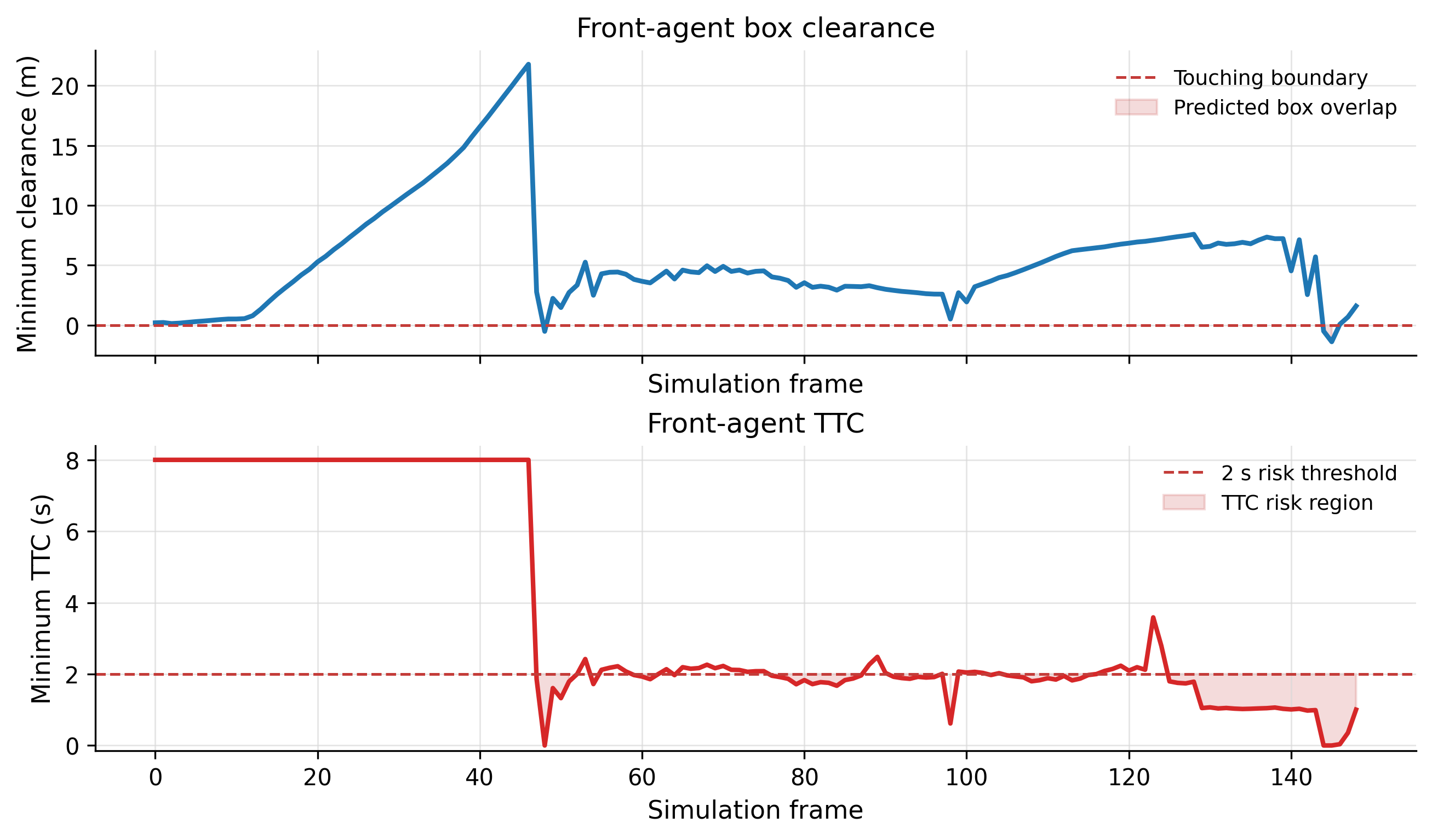}
    \par\vspace{0.3mm}
    \makebox[\linewidth][c]{\scriptsize (b) Front-agent clearance and TTC}
    \par\vspace{1mm}
    \includegraphics[width=\linewidth]{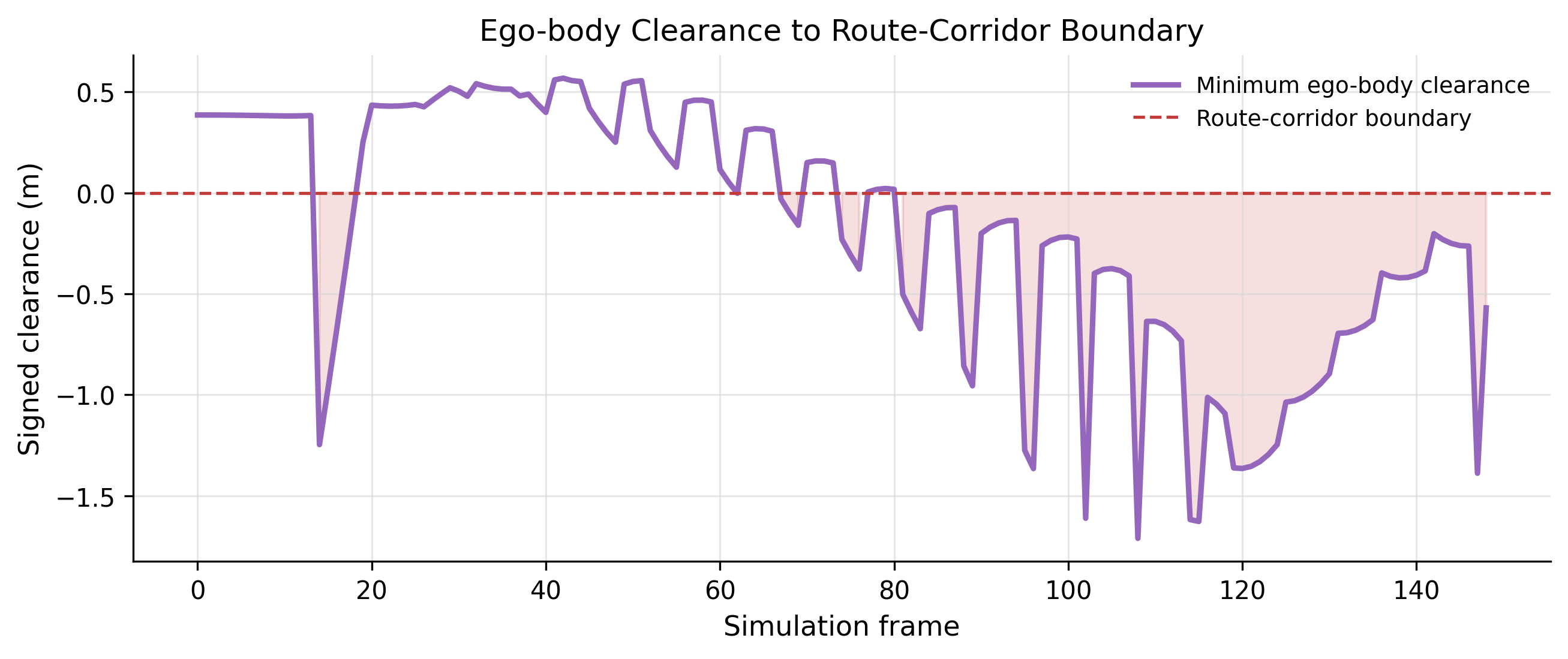}
    \par\vspace{0.3mm}
    \makebox[\linewidth][c]{\scriptsize (c) Signed route-corridor clearance}
\end{minipage}
\caption{A \texttt{starting\_left\_turn} case from test14-hard-r. (a) shows
the Head's frame-wise rule pressures. (b) reports front-agent clearance and TTC. (c)
shows signed ego-body clearance to the route-corridor boundary.}
\label{fig:case-study}
\vspace{-0.4em}
\end{figure}

\begin{figure}[H]
\centering
\begin{minipage}[t]{0.235\textwidth}
    \centering
    \includegraphics[width=\linewidth]{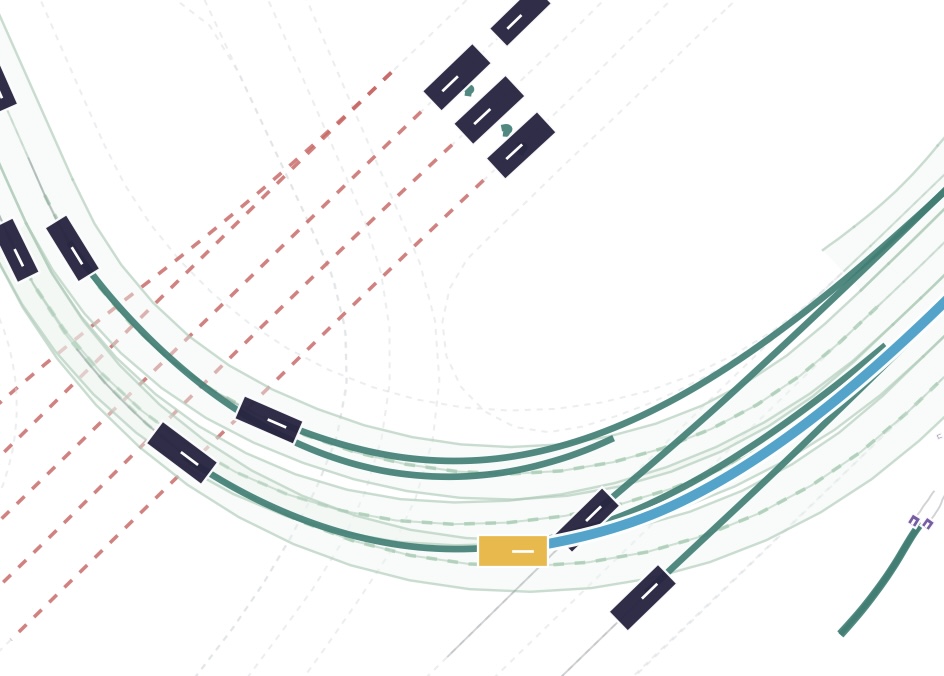}
    \par\smallskip
    \makebox[\linewidth][c]{\small (a) Baseline, frame 100}
\end{minipage}\hfill
\begin{minipage}[t]{0.235\textwidth}
    \centering
    \includegraphics[width=\linewidth]{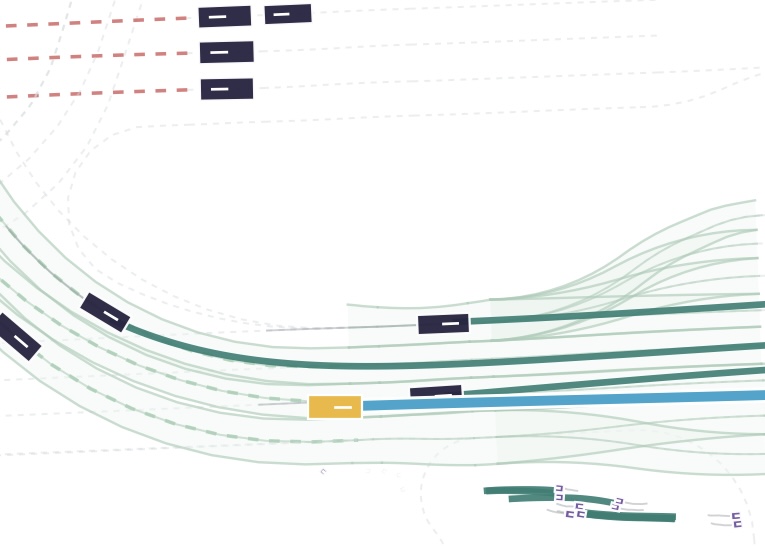}
    \par\smallskip
    \makebox[\linewidth][c]{\small (b) Baseline, frame 140}
\end{minipage}\hfill
\begin{minipage}[t]{0.235\textwidth}
    \centering
    \includegraphics[width=\linewidth]{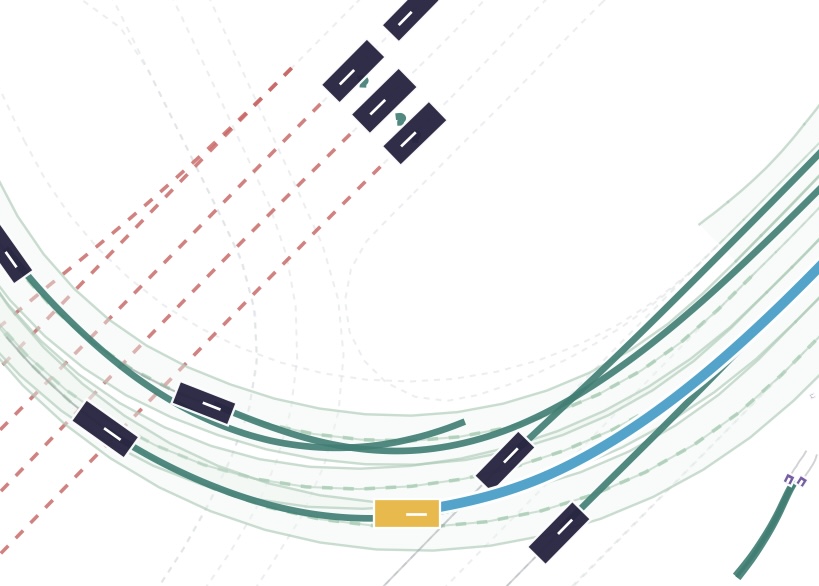}
    \par\smallskip
    \makebox[\linewidth][c]{\small (c) RADP, frame 100}
\end{minipage}\hfill
\begin{minipage}[t]{0.235\textwidth}
    \centering
    \includegraphics[width=\linewidth]{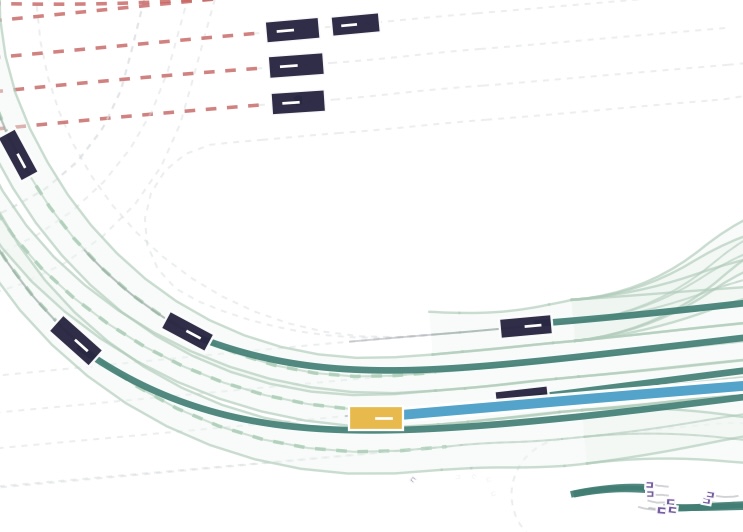}
    \par\smallskip
    \makebox[\linewidth][c]{\small (d) RADP, frame 140}
\end{minipage}
\caption{Matched closed-loop views. Baseline and RADP are shown near the two
critical interactions; yellow and blue denote the ego vehicle and neighbors.}
\label{fig:case-frames}
\vspace{-0.5em}
\end{figure}

\paragraph{\textbf{Left-turn scenario.}}
The rollout contains two collision-critical interactions near frames 100 and
140, where front-agent clearance and TTC decrease
(\radpfigref{fig:case-study}\textcolor{refred}{(b)}); after frame 80, the signed
route-corridor clearance also falls and becomes negative
(\radpfigref{fig:case-study}\textcolor{refred}{(c)}).
Given these observed risks, the Head correctly raises collision pressure around
both interactions and increases lane pressure as the boundary is approached,
while keeping speed, kinematic, and comfort pressures comparatively low
(\radpfigref{fig:case-study}\textcolor{refred}{(a)}). The matched snapshots further show that RADP
maintains greater separation than the baseline (\radpfigref{fig:case-frames}),
providing a qualitative counterpart to \radptabref{tab:fidelity}
and~\radptabref{tab:temporal-val}.

\vspace{-0.8em}
\enlargethispage{10pt}

\subsection{Ablation Study}

Rule supervision improves planning and safety over the rule-free and
unnormalized variants. Collision-only inference guidance performs substantially
worse, supporting training-time rule integration (\radptabref{tab:rule-ablation}).

\begin{table}[H]
\centering
\caption{Rule ablations on test14-hard-r (\(\uparrow\)). Training variants use
the same configuration; collision guidance is applied during DPM sampling.}
\label{tab:rule-ablation}
\scriptsize
\setlength{\tabcolsep}{3.4pt}
\renewcommand{\arraystretch}{1.04}
\resizebox{\columnwidth}{!}{%
\begin{tabular}{lccccccc}
\toprule
\textbf{Setting} & \textbf{Score} & \textbf{Collision} & \textbf{TTC} & \textbf{Drivable} & \textbf{Progress} & \textbf{Speed} & \textbf{Comfort} \\
\midrule
Collision-Guided Diffusion Planner & 59.92 & 76.65 & 68.75 & 90.44 & \best{95.96} & 96.00 & 60.66 \\
\addlinespace[1pt]
FFN-LoRA (no rules) & 69.20 & 88.42 & 81.62 & \best{95.96} & 90.44 & \best{98.37} & \best{86.03} \\
No normalization & 67.78 & 85.11 & 77.57 & 93.75 & 93.38 & 95.93 & 84.56 \\
RADP & \best{70.62} & \best{92.10} & \best{83.46} & 95.59 & 90.07 & 98.29 & 85.29 \\
\bottomrule
\end{tabular}%
}
\renewcommand{\arraystretch}{1}
\end{table}

\section{Limitations and Conclusion}

RADP internalizes differentiable driving rules during diffusion training and predicts rule-specific optimization pressures online. We use the trajectory-gradient magnitude \(\|\nabla_X J_i\|\), rather than the rule cost \(J_i\), as the attribution target: while \(J_i\) measures the current rule violation, its gradient more directly reflects how strongly the rule would locally modify the generated trajectory. Moreover, the current framework assumes a predefined rule set and fixed rule weights, and thus cannot adapt its rule priorities to changing scenarios or preferences. Comparisons with other attribution families are also non-trivial because their explanations are not directly aligned with our rule-specific risk endpoints. Future work will explore intervention-based validation, adaptive rule weighting, broader rule sets, and principled comparisons with other explanation methods.

\clearpage
\subsection*{AI Use Statement}
We used generative AI tools to aid or polish the writing of the manuscript and
to support retrieval and discovery, such as identifying potentially relevant
related work. All AI-assisted text and retrieved references were reviewed and
verified by the authors.

\subsection*{Ethics Statement}
This work is computational and uses the publicly available nuPlan dataset and
simulation framework. It does not involve human-subject experiments or private
personal data. Because autonomous driving is safety-critical, the simulation
results reported in this paper should not be interpreted as evidence of
real-world deployment readiness. Practical deployment would require extensive
real-world testing, safety validation, and appropriate human oversight.

\subsection*{Reproducibility Statement}
We provide detailed descriptions of the planner architecture, differentiable
driving rules, rule normalization, gradient-based Teacher, online attribution
Head, and training procedure in the main paper and appendix. We also document
the evaluation splits, closed-loop simulation settings, risk endpoints,
attribution metrics, temporal controls, and statistical procedures. Code,
configuration files, trained checkpoints, and evaluation scripts will be
released to facilitate reproducibility.

\clearpage
\bibliography{iclr2026_conference}
\bibliographystyle{iclr2027_conference}

\clearpage
\appendix

\section{Baseline Diffusion Objective}
\label{app:diffusion-objective}

Following Diffusion Planner~\citep{zheng2025diffusionplanner}, the forward
process and clean-trajectory prediction are
\begin{equation}
q(\mathbf X^{(k)}\mid\mathbf X^{(0)})=
\mathcal N\!\left(\sqrt{\bar\alpha_k}\mathbf X^{(0)},
(1-\bar\alpha_k)\mathbf I\right),\qquad
\widehat{\mathbf X}^{(0)}=\mu_\theta(\mathbf X^{(k)},k,S_t).
\end{equation}
The masked reconstruction objective is
\begin{align}
\mathcal L_{\mathrm{diff}}&=\mathcal L_{\mathrm{nbr}}+
2\mathcal L_{\mathrm{ego}},\\
\mathcal L_{\mathrm{ego}}&=\mathbb E\!\left[\frac1H\sum_{h=1}^H
\|\widehat X^{(0)}_{\mathrm{ego},h}-X^{(0)}_{\mathrm{ego},h}\|_2^2\right],\\
\mathcal L_{\mathrm{nbr}}&=\mathbb E\!\left[\frac1{|\mathcal V|}
\sum_{(b,n,h)\in\mathcal V}
\|\widehat X^{(0)}_{b,n,h}-X^{(0)}_{b,n,h}\|_2^2\right],
\end{align}
where $\mathcal V$ excludes padded agents and unavailable future states.

\section{Rule and Risk Definitions}
\label{app:rules}

Planner adaptation, gradient supervision, and Head outputs use the same six
semantic channels listed in \radptabref{tab:driving_rules}:
\begin{equation}
\mathcal R_{\mathrm{train}}=\mathcal R_{\mathrm{attr}}
=\{\mathrm{collision},\mathrm{lane},\mathrm{overspeed},
\mathrm{kinematics},\mathrm{comfort},\mathrm{goal}\}.
\end{equation}
Thus, the planner rule loss, Teacher targets, and Head outputs are fully aligned
in semantics and dimensionality. Underspeed and traffic-light costs do not form independent training or
attribution channels. TTC is likewise not an attribution channel: the TTC row
uses collision pressure as its score and evaluates the same attribution
against a distinct future-risk endpoint.

\subsection{Trajectory Denormalization and Finite Differences}

Rules operate on denormalized trajectories at $0.1\,$s resolution. Predicted
states represent rear-axle position and heading; collision computation further
uses the nuPlan Pacifica length, width, and rear-axle-to-center offset to
construct oriented boxes. Speed, acceleration, jerk, curvature, and curvature
rate are obtained by finite differences over the current ego state and future
trajectory. Invalid map points, padded agents, and unavailable future states
are excluded by masks.

\subsection{Complete Instantiation of Rule Costs}

\paragraph{Collision.}
In addition to Eq.~\eqref{eq:collision_violation}, the implementation uses
$t_{\mathrm{head}}=0.8\,$s and
$a_{\mathrm{brake}}=4.0\,\mathrm{m/s^2}$. For each future time and obstacle,
\begin{equation}
\tau_{h,n}=\frac{[d^{\mathrm{box}}_{h,n}]_+}
{\max(c_{h,n},10^{-3})}.
\end{equation}
A stop-gradient soft gate combines box separation, closing speed, a $4\,$s TTC
threshold, and the safety-distance deficit, with a hard cutoff beyond $20\,$m.
Active terms are combined by a softmax-weighted mean with temperature $8$, so
that the most critical interactions receive greater weight.

\paragraph{Lane.}
Each predicted point is matched to the nearest route-polyline point. Its
tangent defines a normal direction and signed lateral error $e_h$. If the
available left and right widths are $w_h^L$ and $w_h^R$, respectively,
\begin{align}
z_h^L&=e_h+W_{\mathrm{ego}}/2-w_h^L,\\
z_h^R&=-e_h+W_{\mathrm{ego}}/2-w_h^R,\\
J_{\mathrm{lane}}&=J(z^L;0.5)+J(z^R;0.5)
+0.05J(|e|;2.0),
\end{align}
where $J(z;\sigma)$ denotes Eq.~\eqref{eq:smooth_violation} averaged over
valid points. The final term is a weak centerline regularizer and does not
replace the ego-body boundary constraints.

\paragraph{Speed.}
Let $v_h^{\lim}$ be the statutory speed limit of the nearest valid route lane.
The current six-dimensional planner and attribution implementation uses only
the atomic overspeed cost,
\begin{equation}
J_{\mathrm{over}}=J(v-v^{\lim};1.0).
\end{equation}
Time steps without a valid route-lane speed limit are masked out. Road
curvature, low-speed deviation, and free-road status do not enter this channel,
preserving its semantics as pressure from exceeding the statutory speed limit.

\paragraph{Kinematics and comfort.}
The kinematic cost is
\begin{align}
J_{\mathrm{kin}}={}&J(a_{\mathrm{lon}}-6;1)+J(-a_{\mathrm{lon}}-8;1)\\
&+J(|a_{\mathrm{lat}}|-4.5;1)+J(|\kappa|-0.35;0.05),
\end{align}
where acceleration is measured in $\mathrm{m/s^2}$ and curvature in
$\mathrm{m^{-1}}$. The comfort cost is
\begin{equation}
J_{\mathrm{comf}}=J(|j_{\mathrm{lon}}|-8.37;1)
+J(|j_{\mathrm{lat}}|-8.37;1)
+J(|\dot\kappa|-0.30;0.1).
\end{equation}

\paragraph{Goal progress.}
Let $s_h$ denote cumulative progress after projecting the trajectory onto the
route. The target $s^\star$ is determined by the valid expert-future endpoint
when available; otherwise, reachable progress is computed from the current
speed and a nominal acceleration of $0.5\,\mathrm{m/s^2}$ and clipped at the
route endpoint. When a red light ahead is active, the target is capped at
$2\,$m before the stop line. Traffic-light state therefore constrains a
reasonable progress target but does not define a seventh attribution channel.
The cost is
\begin{equation}
J_{\mathrm{goal}}=J(s^\star-s_H;5.0)
+0.1J(d_H^{\mathrm{route}};0.5)
+0.2J(s_{h-1}-s_h;5.0).
\end{equation}

\subsection{Rule Constants and Aggregation}

All six differentiable rules use fixed physical thresholds and calibration
constants, which are collected in \radptabref{tab:rule-constants}.

\begin{table*}[t]
\centering
\caption{Principal rule constants used in the implementation.}
\label{tab:rule-constants}
\scriptsize
\setlength{\tabcolsep}{3pt}
\renewcommand{\arraystretch}{1.18}
\begin{tabular}{p{0.09\textwidth}p{0.28\textwidth}p{0.13\textwidth}p{0.08\textwidth}p{0.30\textwidth}}
\toprule
\textbf{Rule} & \textbf{Threshold} & \textbf{Physical scale $\sigma$} & \textbf{Weight $w_i$} & \textbf{Other constants} \\
\midrule
Collision & $d_0=0.5$ m, $t_{\rm head}=0.8$ s, $a_{\rm brake}=4.0$ m/s$^2$ & $0.5$ m & $2.5$ & Active distance $20$ m, risk TTC $4$ s, softmax $\alpha=8$ \\
Lane & Ego-body boundary crossing & $0.5$ m & $0.8$ & Centerline weight $0.05$, center scale $2.0$ m \\
Speed & $v>v^{\lim}$ & $1.0$ m/s & $0.8$ & Time steps without a valid speed limit are masked out \\
Kinematics & $a_{\rm lon}\le6$, $-a_{\rm lon}\le8$, $|a_{\rm lat}|\le4.5$, $|\kappa|\le0.35$ & $1,1,1,0.05$ & $0.5$ & --- \\
Comfort & $|j_{\rm lon}|,|j_{\rm lat}|\le8.37$, $|\dot\kappa|\le0.30$ & $1,1,0.1$ & $0.8$ & --- \\
Goal progress & Reachable/expert progress deficit & $5.0$ m & $1.5$ & Lateral weight $0.1$, monotonicity weight $0.2$, stopping buffer $2$ m \\
\bottomrule
\end{tabular}
\renewcommand{\arraystretch}{1}
\end{table*}

Except for collision, costs are averaged over valid active elements; collision
uses the softmax-weighted mean above. During planner training, a stop-gradient
EMA scale $s_i^{(r)}$ is maintained for each complete rule cost with decay
$0.99$ and lower bound $10^{-3}$, and
$\widetilde J_i=J_i/\operatorname{sg}(s_i^{(r)})$. Teacher targets are always
computed from gradients of atomic costs $J_i$ before semantic weighting or EMA
normalization.

\subsection{External Rollout-Risk Endpoints}

The $v_i^{\mathrm{rollout}}$ in Eq.~\eqref{eq:future_risk} is constructed from
executed ego and agent states in the same closed-loop simulation. It is not the
differentiable rule cost and is not used to train the head. Specifically:
(i) physical collision is a positive-area overlap between ego and tracked-agent
boxes; (ii) TTC risk uses the reciprocal of nuPlan's per-frame TTC; (iii) lane
risk uses the signed ego-body margin to the route/drivable boundary, with
$0.30\,$m or less treated as an event; (iv) overspeed uses the positive part of
the nuPlan speed-limit exceedance; (v) kinematics uses normalized violations of
longitudinal/lateral acceleration and curvature; (vi) comfort uses the maximum
ratio of longitudinal jerk, lateral jerk, and curvature rate to their
thresholds; and (vii) goal risk uses incomplete reference-route progress, with
an event threshold at an ego-to-expert progress ratio below $0.80$.

For planning frame $t_k$, collision and TTC use a $2\,$s future window, while
lane, speed, kinematics, comfort and goal use $8\,$s. Severity in each window is
aggregated by a pre-specified rule-specific operator. Temporal Spearman is
computed within each scenario having sufficient temporal variation and then
averaged across scenarios. The shuffled control permutes attribution only
within the same scenario, preserving its marginal distribution. Confidence
intervals use $10{,}000$ scenario-level bootstrap resamples.

\subsection{Complete Head--Teacher Fidelity Metrics}
\label{app:fidelity-metrics}

Let $\mathcal F$ be the valid synchronized frames and $M$ the number of rules.
We compute per-rule correlation and its macro-average as
\begin{equation}
\rho_i^{H,T}=\operatorname{Spearman}
\left(\{A_i^H(t)\}_{t\in\mathcal F},\{A_i^T(t)\}_{t\in\mathcal F}\right),
\qquad
\rho_{\mathrm{macro}}^{H,T}=\frac1M\sum_{i=1}^M\rho_i^{H,T}.
\label{eq:fidelity_spearman}
\end{equation}
Numerical agreement and dominant-rule agreement are
\begin{align}
\operatorname{MAE}^{H,T}
&=\frac{1}{M|\mathcal F|}\sum_{i=1}^M\sum_{t\in\mathcal F}
|A_i^H(t)-A_i^T(t)|,\label{eq:fidelity_mae}\\
\operatorname{Top\text{-}1}^{H,T}
&=\frac{1}{|\mathcal F|}\sum_{t\in\mathcal F}
\mathbf 1\!\left[\arg\max_i A_i^H(t)=\arg\max_i A_i^T(t)\right].
\label{eq:fidelity_top1}
\end{align}

\subsection{Complete Temporal-Alignment Metrics}
\label{app:temporal-metrics}

For rule $i$ and scenario $n$, continuous temporal alignment is
\begin{equation}
\rho_{i,n}^{\mathrm{temp}}=\operatorname{Spearman}
\!\left(\{A_{i,n}(t_k)\}_k,\{R_{i,n}(t_k)\}_k\right),\qquad
\rho_i^{\mathrm{temp}}=\frac{1}{|\mathcal N_i^\rho|}
\sum_{n\in\mathcal N_i^\rho}\rho_{i,n}^{\mathrm{temp}},
\end{equation}
where $\mathcal N_i^\rho$ contains scenarios with enough valid frames and
non-constant attribution and risk sequences. For attribution source
$Q\in\{T,H\}$, the reported macro score averages all seven evaluated endpoints,
\begin{equation}
\rho_{\mathrm{macro}}^{Q}
=\frac{1}{7}\sum_{e\in\mathcal E}\rho_e^{Q},\quad
\mathcal E=\{\text{collision, TTC, lane, speed, kinematics, comfort, goal}\}.
\end{equation}
Physical collision and TTC are separate endpoints that share the collision
attribution channel. With a pre-specified threshold,
$y_{i,n}(t_k)=\mathbf 1[R_{i,n}(t_k)>\delta_i]$. AUPRC uses attribution as
the continuous event-retrieval score and is macro-averaged over eligible
scenarios. If $\mathcal T_{i,n}^{10}$ is the top 10\% of valid frames ranked by
attribution, then
\begin{equation}
\operatorname{Lift@10\%}_{i,n}=
\frac{|\mathcal T_{i,n}^{10}|^{-1}\sum_{t_k\in\mathcal T_{i,n}^{10}}y_{i,n}(t_k)}
{|\mathcal T_{i,n}|^{-1}\sum_{t_k\in\mathcal T_{i,n}}y_{i,n}(t_k)},
\qquad
\operatorname{Lift@10\%}_{i}=\frac{1}{|\mathcal N_i^L|}
\sum_{n\in\mathcal N_i^L}\operatorname{Lift@10\%}_{i,n}.
\end{equation}
The random-ranking AUPRC baseline equals the positive-event rate. The temporal
control permutes attribution within each scenario before recomputing Spearman.
All 95\% confidence intervals use $10{,}000$ scenario-level bootstrap
resamples, retaining all correlated frames of each sampled scenario.

\section{Additional Experimental Details}

\subsection{Non-Reactive Planning Results}

For completeness, \radptabref{tab:planner-nr} reports the corresponding
non-reactive aggregate scores. These results are supplementary to the reactive
evaluation emphasized in the main text.

\begin{table}[h]
\centering
\caption{Closed-loop planning scores with non-reactive agents ($\uparrow$).}
\label{tab:planner-nr}
\small
\begin{tabular}{lcc}
\toprule
\textbf{Split} & \textbf{Diffusion Planner} & \textbf{RADP} \\
\midrule
val14 & \best{89.65} & 88.72 \\
test14-random & 88.88 & \best{89.46} \\
test14-hard & \best{75.01} & 74.80 \\
\bottomrule
\end{tabular}
\end{table}

\subsection{Planner Adaptation}

We initialize from the released Diffusion Planner checkpoint. The DiT contains
three blocks, each ordered as self-attention, \texttt{mlp1}, scene-conditioned
cross-attention, and \texttt{mlp2}. All baseline parameters are frozen, and
LoRA is inserted only into
\begin{equation}
\{\texttt{mlp1.fc1},\texttt{mlp1.fc2},
\texttt{mlp2.fc1},\texttt{mlp2.fc2}\}
\end{equation}
of every block. For original linear layer $W_0$,
\begin{equation}
W_{\mathrm{eff}}x=W_0x+\frac{\alpha}{r}BAx,
\qquad r=8,\quad\alpha=16,
\end{equation}
with LoRA dropout $0.05$. Twelve linear modules are replaced, yielding 24
trainable $A/B$ tensors; attention, AdaLN, the encoder, input projection, and
final trajectory layer remain frozen. Adaptation uses batch size $128$,
learning rate $2\times10^{-5}$, and random seed $3407$.

\subsection{Teacher Calibration and Caching}

To make gradients with different physical dimensions comparable, we collect
RMS gradients for the six attribution channels over 100 calibration batches from
the training split and use the 75th percentile of each channel's non-negative
gradients as $\kappa_i$. Equation~\eqref{eq:teacher} then produces
log-compressed absolute pressures without normalization across the rule
dimension.

With the planner frozen, caching records the final-DPM denoising state used in
closed-loop inference: final DiT agent tokens, ego-trajectory geometry, route
polylines, neighbor masks, and teacher pressures. The training cache contains
$100{,}000$ samples from \texttt{processed}; the validation cache contains
$20{,}000$ non-overlapping samples from \texttt{new\_processed}.

\subsection{Trajectory-Conditioned Scene Attention Head Details}

The attribution head operates on the final denoising representation produced by the frozen rule-aligned planner. The token dimension is 192. For each planning frame, we construct a trajectory-conditioned query \(z_t\) by combining the final-DPM ego token with geometric features extracted from the generated ego trajectory. The scene-condition representation consists of the valid context tokens provided by the planner encoder, covering surrounding agents, map elements, route information, and lane geometry. All planner-side representations are detached and remain frozen during attribution-head training.

To recover rule-relevant scene context, \(z_t\) attends to the scene-condition tokens through a six-head multi-head attention module,

$$ c_t=\operatorname{MHA}_{\mathrm{scene}}(z_t,S_t,S_t), $$

where \(S_t\) denotes the valid scene-condition token set and the corresponding validity masks are applied during attention. This trajectory-conditioned attention allows the same generated trajectory to retrieve different contextual evidence depending on the surrounding traffic and road geometry, rather than aggregating agent and route information through separate attention branches.

The query representation \(z_t\) and attended scene context \(c_t\) are concatenated and passed to six rule-specific prediction branches, corresponding to collision, lane, overspeed, kinematics, comfort, and goal progress. Each branch uses an independent two-layer MLP with hidden dimension 256 and an unshared scalar output layer. Softplus is applied to each scalar output to obtain a non-negative rule pressure. Thus, the six branches share the trajectory and scene representations but do not share their final prediction parameters.

The attribution head is trained from scratch for 50 epochs with batch size 512 using AdamW, a learning rate of \(10^{-4}\), and weight decay \(10^{-4}\). The Smooth-L1 regression loss uses transition parameter \(\beta_{\mathrm{Huber}}=0.2\). To improve learning in the high-pressure tail, collision samples above the 75th percentile of positive collision pressure are sampled with a \(3.0\times\) multiplier and assigned a regression weight of \(2.0\). Lane samples above the 85th percentile are sampled with a \(1.5\times\) multiplier. For the remaining rule channels, samples above the corresponding 85th-percentile pressure threshold are sampled with a \(1.5\times\) multiplier and assigned a loss weight of \(1.25\). The pairwise-ranking loss is weighted by \(0.2\), uses temperature \(0.2\), and ignores teacher-pressure pairs whose absolute difference is below \(0.05\).

Throughout attribution-head training, all planner parameters, including the LoRA adapters, remain frozen. The head is supervised exclusively by the calibrated gradient-pressure Teacher targets cached from the frozen planner; future closed-loop rollout risks are never used as training supervision.

\subsection{Evaluation Reproducibility}

Teacher and head use the same planner checkpoint, scene state, and generated
trajectory at each frame. Head--teacher fidelity is computed directly on these
synchronized frames. For behavioral alignment, $A_i(t_k)$ is paired with
future risk over $[t_k,t_k+H_i]$ in the same rollout. The random AUPRC baseline
equals the positive-event rate, and Lift@10\% uses the highest-attribution 10\%
of all valid frames. All thresholds, windows, and aggregation operators are
fixed before inspecting head results and shared by teacher and head.

\subsection{Latency Details}
\label{app:latency}

End-to-end latency is measured over five fixed test14-hard-r scenarios, while
cached-state attribution is measured over 500 synchronized frames. The former
includes trajectory generation, whereas the latter isolates attribution
computation; all measurements use the same RTX~4080. The complete mean and
standard-deviation results are reported in \radptabref{tab:latency-details}.

\begin{table}[h]
\centering
\caption{Inference latency details on an RTX~4080. Values are mean $\pm$
standard deviation; lower is better.}
\label{tab:latency-details}
\small
\begin{tabular}{lrr}
\toprule
\textbf{Method} & \textbf{Latency (ms)} & \textbf{Overhead} \\
\midrule
Planner only & $251.05\pm39.04$ & --- \\
Planner + Head & $245.50\pm30.43$ & $-1.64\%\pm6.67\%$ \\
Planner + exact Teacher & $465.29\pm66.46$ & $+89.18\%\pm38.35\%$ \\
Planner + gradient guidance & $571.10\pm52.64$ & $+130.76\%\pm30.01\%$ \\
\midrule
Exact Teacher on cached states & $62.51\pm3.69$ & $1.00\times$ \\
Head on cached states & \best{$8.18\pm1.20$} & \best{$7.65\times$ faster} \\
\bottomrule
\end{tabular}
\end{table}

\subsection{Complete Confidence Intervals for Main-Text Attribution Results}
\label{app:attribution-ci}

The scenario-bootstrap 95\% confidence intervals omitted from the main text for
readability are reported in \radptabref{tab:fidelity-ci} and
\radptabref{tab:temporal-val-ci}. Resampling is performed at the scenario level so
that correlated frames from the same scenario are not treated as independent
observations.

\begin{table*}[t]
\centering
\caption{95\% confidence intervals for Head--Teacher fidelity and macro-averaged future-risk alignment on reactive splits.}
\label{tab:fidelity-ci}
\footnotesize
\renewcommand{\arraystretch}{1.15}
\setlength{\tabcolsep}{3.5pt}
\resizebox{\textwidth}{!}{%
\begin{tabular}{lccccc}
\toprule
\multirow{2}{*}{\textbf{Reactive split}} & \multicolumn{3}{c}{\textbf{Head--Teacher fidelity 95\% CI}} & \multicolumn{2}{c}{\textbf{Future-risk alignment 95\% CI}} \\
\cmidrule(lr){2-4}\cmidrule(lr){5-6}
& \textbf{Macro Spearman} & \textbf{MAE} & \textbf{Top-1} & \textbf{Teacher $\rho_{\mathrm{macro}}$} & \textbf{Head $\rho_{\mathrm{macro}}$} \\
\midrule
val14 & [0.525, 0.547] & [0.209, 0.222] & [69.4, 71.8]\% & [0.213, 0.256] & [0.183, 0.229] \\
test14-random & [0.552, 0.588] & [0.188, 0.212] & [69.5, 73.9]\% & [0.133, 0.258] & [0.119, 0.233] \\
test14-hard & [0.537, 0.571] & [0.224, 0.251] & [69.7, 74.0]\% & [0.184, 0.255] & [0.151, 0.231] \\
\bottomrule
\end{tabular}}
\renewcommand{\arraystretch}{1}
\end{table*}

\begin{table*}[t]
\centering
\caption{Scenario-bootstrap 95\% confidence intervals for rule-wise future-risk alignment on val14-r.}
\label{tab:temporal-val-ci}
\footnotesize
\renewcommand{\arraystretch}{1.15}
\setlength{\tabcolsep}{3.5pt}
\resizebox{\textwidth}{!}{%
\begin{tabular}{llcccccc}
\toprule
\textbf{Attribution channel} & \textbf{Risk endpoint} & \textbf{Teacher AUPRC} & \textbf{Head AUPRC} & \textbf{Teacher Lift@10\%} & \textbf{Head Lift@10\%} & \textbf{Teacher $\rho$} & \textbf{Head $\rho$} \\
\midrule
Collision & Physical collision & [0.481,0.604] & [0.452,0.578] & [1.388,2.257] & [1.184,2.019] & [0.158,0.302] & [0.077,0.252] \\
Collision & TTC risk & [0.516,0.598] & [0.450,0.527] & [1.640,2.212] & [1.334,1.896] & [0.196,0.274] & [0.115,0.198] \\
Lane & Lane-boundary risk & [0.894,0.928] & [0.815,0.864] & [1.422,1.617] & [1.260,1.428] & [0.283,0.345] & [0.206,0.265] \\
Speed & Overspeed risk & [0.725,0.829] & [0.701,0.816] & [1.553,2.254] & [1.556,2.285] & [0.198,0.402] & [0.200,0.388] \\
Kinematics & Kinematic risk & [0.167,0.167] & [0.250,0.250] & [0.000,0.000] & [0.000,0.000] & [0.079,0.143] & [0.123,0.194] \\
Comfort & Comfort risk & [0.162,0.679] & [0.156,0.598] & [0.000,4.375] & [0.000,3.500] & [0.072,0.133] & [0.089,0.155] \\
Goal progress & Route-progress deficit & [0.829,0.871] & [0.774,0.822] & [1.687,1.924] & [1.513,1.740] & [0.314,0.370] & [0.286,0.345] \\
\bottomrule
\end{tabular}}
\renewcommand{\arraystretch}{1}
\end{table*}

\clearpage
\section{Additional Collision-Avoidance Cases}
\label{app:collision-cases}

We further compare matched Baseline and RADP rollouts on four test14-hard-r
scenarios for which the Baseline incurs an at-fault collision while RADP does
not.  For each scenario, the upper row shows the Baseline and the lower row
shows RADP at exactly the same simulation indices.  The four columns correspond
to approximately three, two, and one second before the Baseline's first
geometric contact, followed by the contact frame itself.  Yellow boxes denote
the ego vehicle, dark boxes denote other traffic participants, cyan curves show
the trajectory planned at the displayed frame, and orange curves show the
executed ego history.  At the contact frame, the colliding participant and the
Baseline ego outline are highlighted in red.  All panels use the original
nuPlan vector-map geometry and a fixed world-coordinate viewport within each
scenario.

\begin{figure*}[h!]
    \centering
    \includegraphics[width=0.97\textwidth]{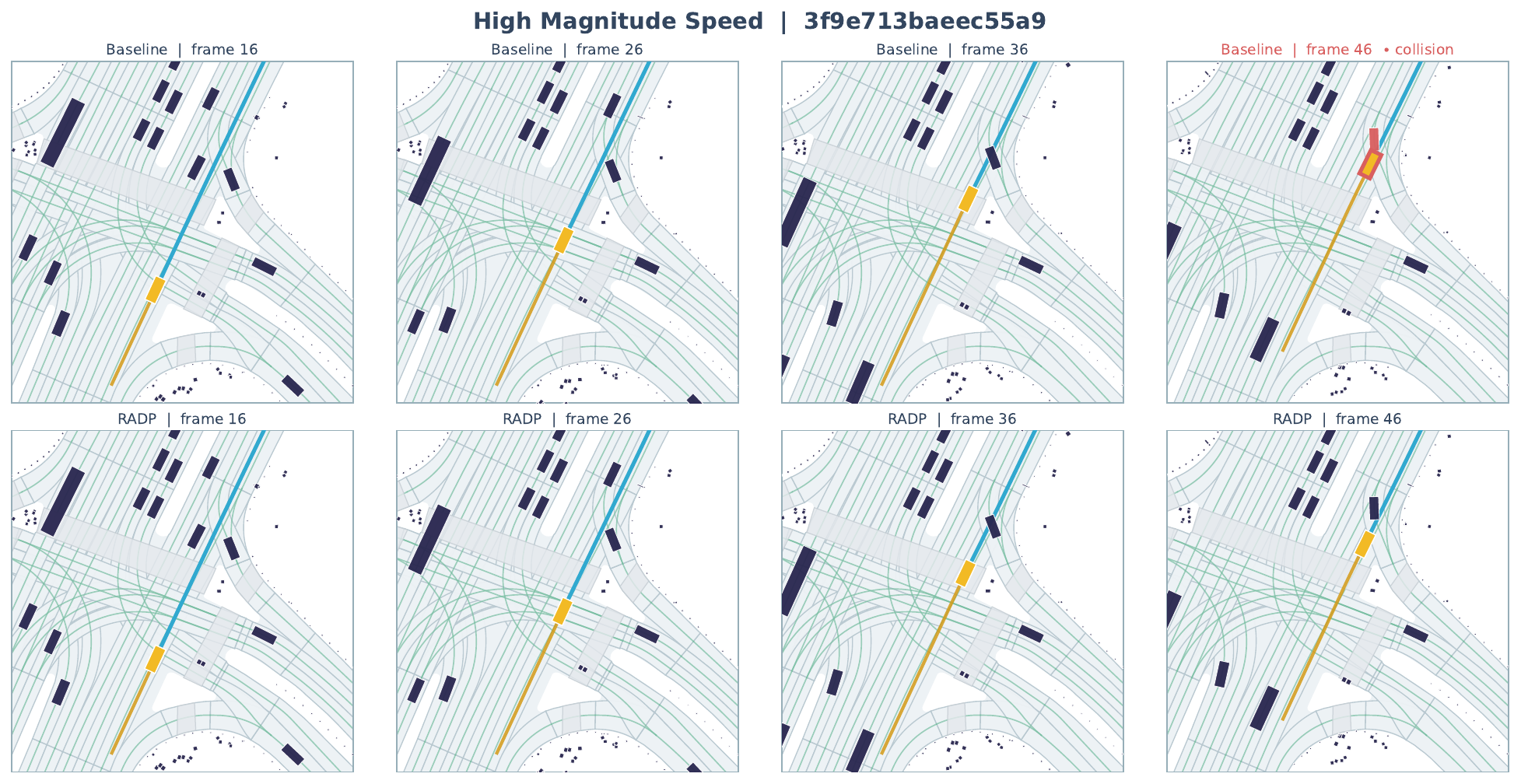}
    \vspace{2mm}
    \includegraphics[width=0.97\textwidth]{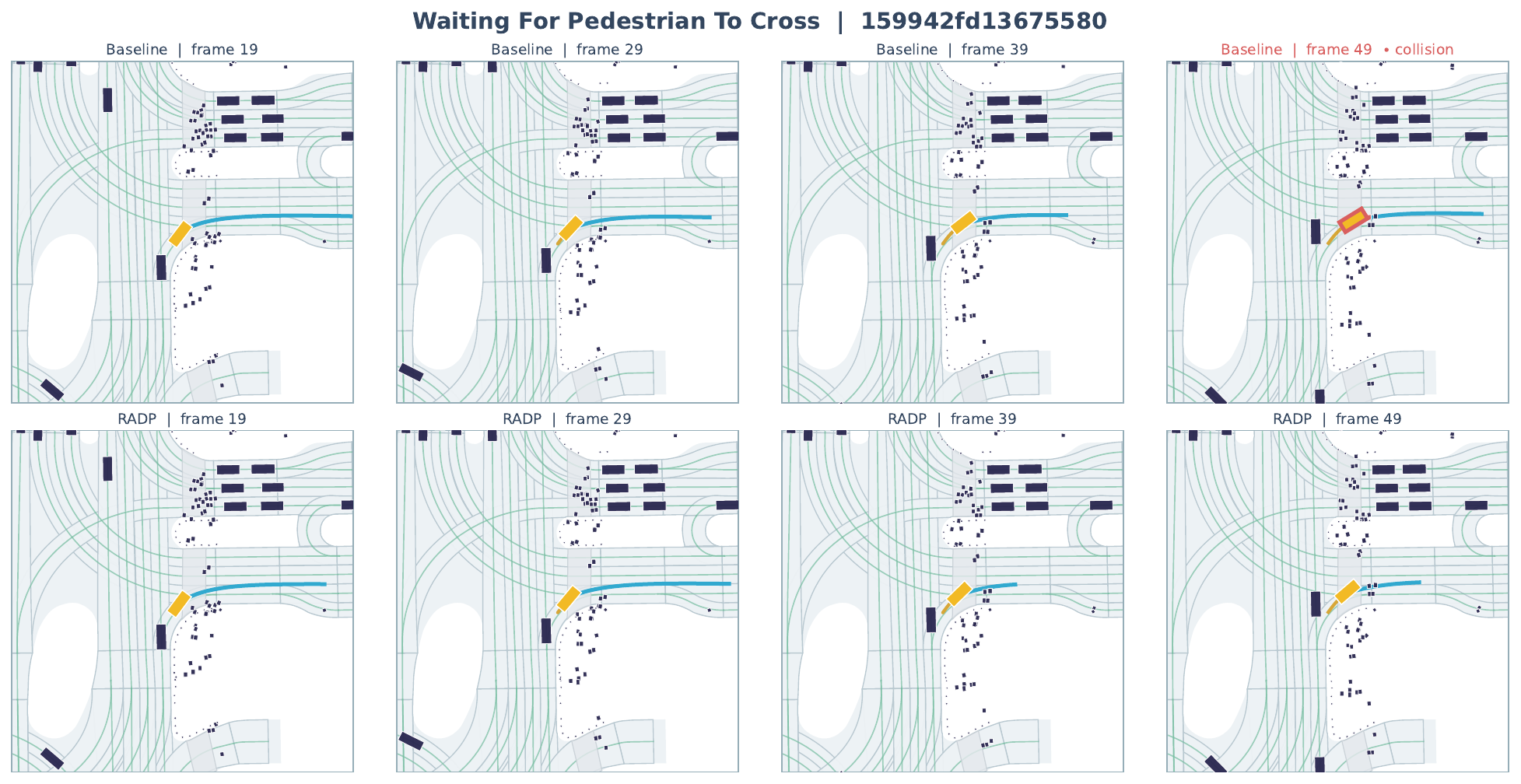}
    \caption{Matched collision-avoidance rollouts. \emph{Top:} a high-speed
    vehicle interaction (token \texttt{3f9e713baeec55a9}). \emph{Bottom:} a
    pedestrian-crossing interaction (token \texttt{159942fd13675580}).  The
    Baseline makes contact in the final column, whereas RADP remains
    collision-free at the corresponding frame.}
    \label{fig:appendix-collision-cases-1}
\end{figure*}

\begin{figure*}[p]
    \centering
    \includegraphics[width=0.97\textwidth]{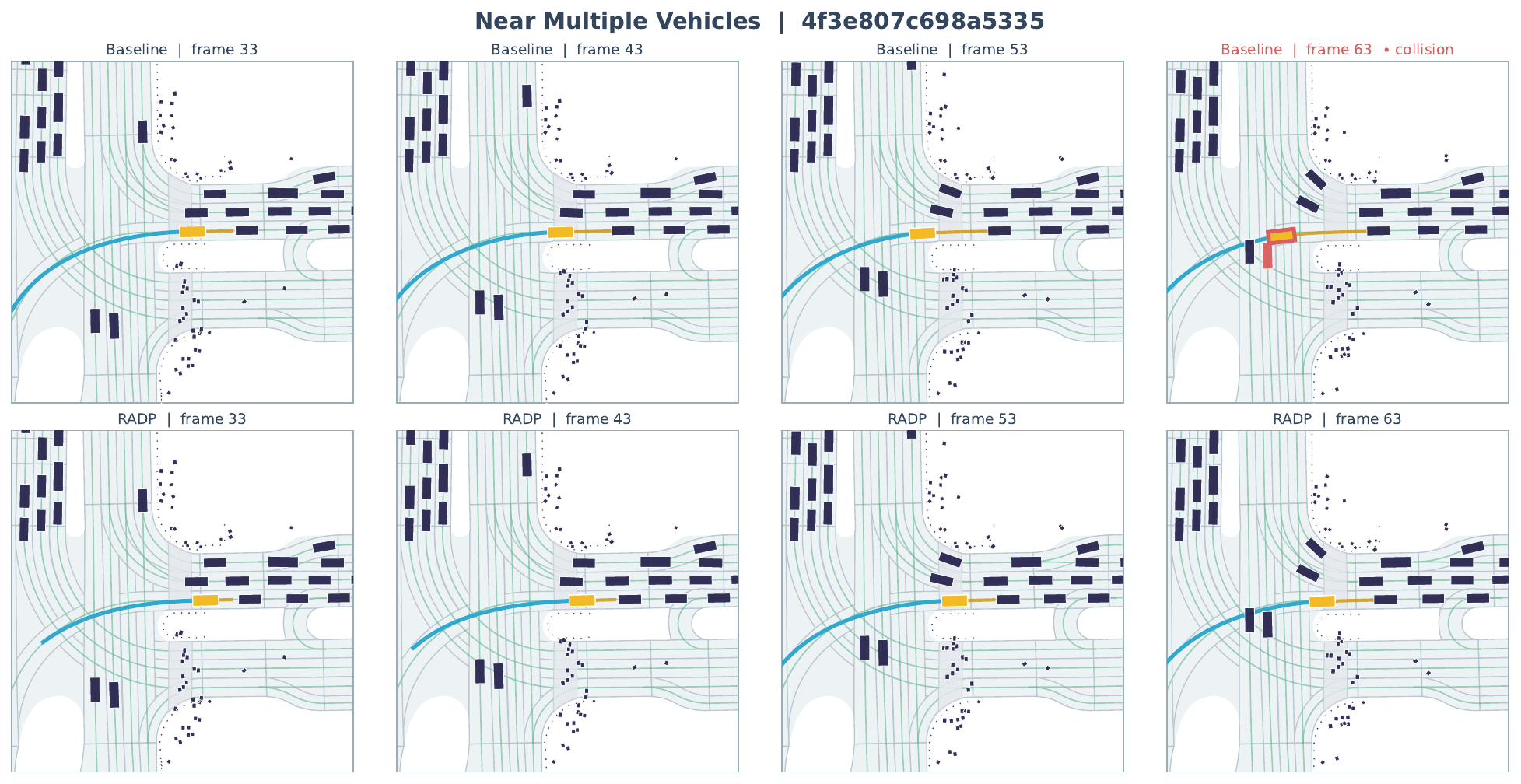}
    \vspace{2mm}
    \includegraphics[width=0.97\textwidth]{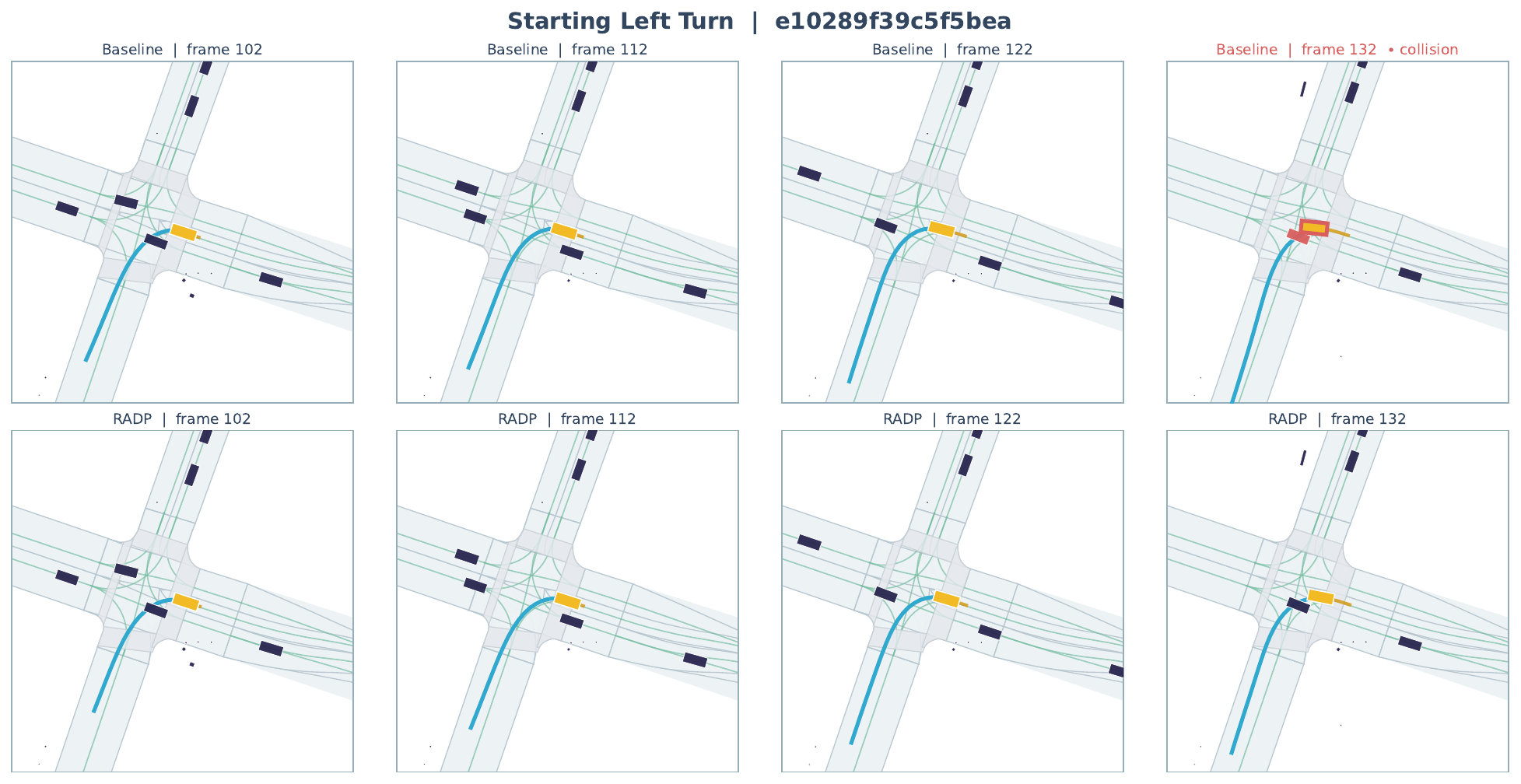}
    \caption{Additional matched collision-avoidance rollouts. \emph{Top:} a
    dense multi-vehicle interaction (token \texttt{4f3e807c698a5335}).
    \emph{Bottom:} a left-turn vehicle conflict (token
    \texttt{e10289f39c5f5bea}).  Identical frame indices are used for Baseline
    and RADP in every column.}
    \label{fig:appendix-collision-cases-2}
\end{figure*}

\end{document}